\documentclass[11pt,letterpaper]{article}
\usepackage[dvipsnames]{xcolor}

\usepackage{emnlp2017}
\usepackage{times}
\usepackage{latexsym}
\usepackage{amsmath}
\usepackage{amssymb}
\usepackage{graphicx}
\usepackage{ifthen}
\usepackage[utf8]{inputenc}
\usepackage{url}

\usepackage{framed}
\usepackage{color}
\usepackage{xspace}
\usepackage{algorithmicx}
\usepackage{algpseudocode}
\usepackage{pstricks}
\usepackage{multirow}
\usepackage{multicol}

\emnlpfinalcopy

\def\emnlppaperid{***}

\newcommand\eqdef{\ensuremath{\stackrel{\rm def}{=}}} % Equal by definition
\ifthenelse{\isundefined{\definition}}{\newtheorem{definition}{Definition}}{}
\ifthenelse{\isundefined{\assumption}}{\newtheorem{assumption}{Assumption}}{}
\ifthenelse{\isundefined{\hypothesis}}{\newtheorem{hypothesis}{Hypothesis}}{}
\ifthenelse{\isundefined{\proposition}}{\newtheorem{proposition}{Proposition}}{}
\ifthenelse{\isundefined{\theorem}}{\newtheorem{theorem}{Theorem}}{}
\ifthenelse{\isundefined{\lemma}}{\newtheorem{lemma}{Lemma}}{}
\ifthenelse{\isundefined{\corollary}}{\newtheorem{corollary}{Corollary}}{}
\ifthenelse{\isundefined{\alg}}{\newtheorem{alg}{Algorithm}}{}
\ifthenelse{\isundefined{\example}}{\newtheorem{example}{Example}}{}
\newcommand\addsent[0]{\mbox{\textsc{AddSent}}\xspace}
\newcommand\addwords[0]{\mbox{\textsc{AddAny}}\xspace}
\newcommand\addonesent[0]{\mbox{\textsc{AddOneSent}}\xspace}
\newcommand\addcommon[0]{\mbox{\textsc{AddCommon}}\xspace}
\newcommand\addmod[0]{\mbox{\textsc{AddSentMod}}\xspace}

\newcommand\nl[1]{``\textit{#1}''}

\definecolor{darkgreen}{rgb}{0,1.0,0}

\title{Adversarial Examples for Evaluating Reading Comprehension Systems}

\author{
  Robin Jia \\
  Computer Science Department \\
  Stanford University \\
  {\tt robinjia@cs.stanford.edu}
\And
	Percy Liang \\
  Computer Science Department \\
  Stanford University \\
  {\tt pliang@cs.stanford.edu}
}

\date{}

\begin{document}

\maketitle

\begin{abstract}
Standard accuracy metrics indicate that 
reading comprehension systems are making rapid progress,
but the extent to which these systems truly understand language remains unclear.
To reward systems with real language understanding abilities,
we propose an adversarial evaluation scheme for the Stanford
Question Answering Dataset (SQuAD). 
Our method tests whether systems can answer questions
about paragraphs that contain adversarially inserted sentences,
which are automatically generated to distract computer systems
without changing the correct answer or misleading humans.
In this adversarial setting,
the accuracy of sixteen published models
drops from an average of $75\%$ F1 score to $36\%$;
when the adversary is allowed to add ungrammatical sequences of words,
average accuracy on four models decreases further to $7\%$.
We hope our insights will motivate
the development of new models that
understand language more precisely.

\end{abstract}

\section{Introduction}
Quantifying the extent to which a computer system exhibits intelligent behavior 
is a longstanding problem in AI \citep{levesque2013best}.
Today, the standard paradigm is to
measure average error across a held-out test set.
However, models can succeed in this paradigm by recognizing patterns
that happen to be predictive on most of the test examples,
while ignoring deeper, more difficult phenomena 
\cite{rimell2009unbounded,paperno2016lambada}.

\begin{figure}[t]
\small
\begin{framed}
\footnotesize
  \textbf{Article:} Super Bowl 50

  \textbf{Paragraph:}
  \nl{Peyton Manning became the first quarterback ever to lead two different teams to multiple Super Bowls. He is also the oldest quarterback ever to play in a Super Bowl at age 39. The past record was held by John Elway, who led the Broncos to victory in Super Bowl XXXIII at age 38 and is currently Denver's Executive Vice President of Football Operations and General Manager.
  \textcolor{blue}{Quarterback Jeff Dean had jersey number 37 in Champ Bowl XXXIV.}}

  \textbf{Question:} \nl{What is the name of the quarterback who was 38 in Super Bowl XXXIII?}

  \textbf{Original Prediction:} \textcolor{Green}{John Elway}

  \textbf{Prediction under adversary:} \textcolor{red}{Jeff Dean}

\end{framed}
\caption{
  An example from the SQuAD dataset.
  The BiDAF Ensemble model originally gets the answer correct,
  but is fooled by the addition of an adversarial
  distracting sentence
  (\textcolor{blue}{in blue}).
}
\label{fig:intro}
\end{figure}

In this work, we propose adversarial evaluation for NLP,
in which systems are instead evaluated on adversarially-chosen inputs.
We focus on the SQuAD reading comprehension task \citep{rajpurkar2016squad},
in which systems answer questions about paragraphs from Wikipedia.
Reading comprehension is an appealing testbed for adversarial evaluation,
as existing models appear successful by 
standard average-case evaluation metrics:
the current state-of-the-art system achieves $84.7\%$ F1 score,
while human performance is just 
$91.2\%$.\footnote{\url{https://rajpurkar.github.io/SQuAD-explorer/}}
Nonetheless, it seems unlikely that existing systems
possess true language understanding
and reasoning capabilities.

Carrying out adversarial evaluation on SQuAD
requires new methods that
adversarially alter reading comprehension examples.
Prior work in computer vision adds imperceptible 
adversarial perturbations to input images,
relying on the fact that such small perturbations
cannot change an image's true label
\citep{szegedy2014intriguing,goodfellow2015explaining}.
In contrast, changing even one word of a paragraph can 
drastically alter its meaning.
Instead of relying on semantics-preserving perturbations,
we create adversarial examples by adding distracting sentences
to the input paragraph, as shown in Figure~\ref{fig:intro}.
We automatically generate these sentences so that they confuse models,
but do not contradict the correct answer or confuse humans.
For our main results,
we use a simple set of rules to generate a raw distractor sentence
that does not answer the question but looks related;
we then fix grammatical errors via crowdsourcing.
While adversarially perturbed images
punish model \emph{oversensitivity} to imperceptible noise,
our adversarial examples target model \emph{overstability}---the
inability of a model to distinguish
a sentence that actually answers the question
from one that merely has words
in common with it.

Our experiments demonstrate that no published open-source model
is robust to the addition of adversarial sentences.
Across sixteen such models,
adding grammatical adversarial sentences
reduces F1 score from an average of $75\%$ to $36\%$.
On a smaller set of four models,
we run additional experiments in which the adversary 
adds non-grammatical sequences of English words,
causing average F1 score to drop further to $7\%$.
To encourage the development of new models that 
understand language more precisely,
we have released all of our code and data publicly.

\section{The SQuAD Task and Models}
\subsection{Task}
The SQuAD dataset \cite{rajpurkar2016squad}
contains 107,785 human-generated reading comprehension questions
about Wikipedia articles.
Each question refers to one paragraph of an article,
and the corresponding answer is guaranteed to be a span in that paragraph.

\subsection{Models}
When developing and testing our methods, we focused
on two published model architectures:
BiDAF \citep{seo2016bidaf} and Match-LSTM \cite{wang2016machine}.
Both are deep learning architectures
that predict a probability distribution over the correct answer.
Each model has a single and an ensemble version,
yielding four systems in total.

We also validate our major findings on twelve other published models
with publicly available test-time code:
ReasoNet Single and Ensemble versions \cite{shen2017reasonet},
Mnemonic Reader Single and Ensemble versions \cite{hu2017mnemonic},
Structural Embedding of Dependency Trees (SEDT) Single and Ensemble versions \cite{liu2017sect},
jNet \cite{zhang2017jnet},
Ruminating Reader \cite{gong2017ruminating},
Multi-Perspective Context Matching (MPCM) Single version \cite{wang2016multi},
RaSOR \cite{lee2017rasor},
Dynamic Chunk Reader (DCR) \cite{yu2016dcr},
and the Logistic Regression Baseline \citep{rajpurkar2016squad}.
We did not run these models during development,
so they serve as a held-out set that validates the generality of our approach.

\subsection{Standard Evaluation}
Given a model $f$ that takes in
paragraph-question pairs $(p, q)$ and outputs an answer $\hat{a}$,
the \emph{standard accuracy} over a test set $D_{\text{test}}$ is simply \[
  \text{Acc}(f) \eqdef \frac1{|D_{\text{test}}|}\sum_{(p, q, a) \in D_{\text{test}}}
  v((p, q, a), f),
\]
where $v$ is the F1 score between the true answer $a$ and the predicted answer $f(p, q)$
(see~\citet{rajpurkar2016squad} for details).

\section{Adversarial Evaluation}
\subsection{General Framework}
A model that relies on superficial cues
without understanding language
can do well according to average F1 score, 
if these cues happen to be predictive most of the time.
\citet{weissenborn2017fastqa} argue that
many SQuAD questions can be answered with heuristics
based on type and keyword-matching.
To determine whether existing models have learned much beyond
such simple patterns, we introduce adversaries that 
confuse deficient models by altering test examples.
Consider the example in Figure~\ref{fig:intro}:
the BiDAF Ensemble model originally gives the right answer,
but gets confused when an adversarial distracting sentence is added
to the paragraph.

We define an adversary $A$ to be a function that
takes in an example $(p, q, a)$, optionally with a model $f$,
and returns a new example $(p', q', a')$.
The \emph{adversarial accuracy} with respect to $A$ is
\[
  \text{Adv}(f) \eqdef \frac1{|D_{\text{test}}|}\sum_{(p, q, a) \in D_{\text{test}}}
           v(A(p, q, a, f), f)).
\]
While standard test error measures the fraction of the test distribution
over which the model gets the correct answer,
the adversarial accuracy measures the fraction over which
the model is \emph{robustly} correct,
even in the face of adversarially-chosen alterations.
For this quantity to be meaningful,
the adversary must satisfy two basic requirements:
first, it should
always generate $(p', q', a')$ tuples that are \emph{valid}---a human
would judge $a'$ as the correct answer to $q'$ given $p'$.
Second, $(p', q', a')$ should be somehow ``close'' to
the original example $(p, q, a)$.

\subsection{Semantics-preserving Adversaries}
In image classification, adversarial examples are commonly generated by adding
an imperceptible amount of noise to the input
\citep{szegedy2014intriguing,goodfellow2015explaining}.
These perturbations do not change the semantics of the image,
but they can change the predictions of models
that are \emph{oversensitive} to semantics-preserving changes.
For language, the direct analogue would be to paraphrase the input
\citep{madnani2010generating}.
However, high-precision paraphrase generation is challenging,
as most edits to a sentence do actually change its meaning.

\subsection{Concatenative Adversaries}
\begin{table}[t]
  \small
  \centering
  \begin{tabular}{|l|c|c|}
    \hline
    & \multicolumn{1}{c|}{Image} & \multicolumn{1}{c|}{Reading} \\
    & \multicolumn{1}{c|}{Classification} & \multicolumn{1}{c|}{Comprehension} \\
    \hline
    \multirow{3}{0.5in}{Possible Input} 
    & \multirow{3}{*}{\begin{minipage}{.15\textwidth}
      \centering
      \includegraphics[scale=0.4]{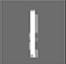}
    \end{minipage}} 
    & \multicolumn{1}{l|}{Tesla moved} \\
    & & \multicolumn{1}{l|}{to the city of} \\
    & & \multicolumn{1}{l|}{Chicago in 1880.} \\
    \hline
    \multirow{3}{0.5in}{Similar Input} 
    & \multirow{3}{*}{\begin{minipage}{.15\textwidth}
      \centering
      \includegraphics[scale=0.4]{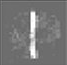}
    \end{minipage}} 
    & \multicolumn{1}{l|}{Tadakatsu moved} \\
    & & \multicolumn{1}{l|}{to the city of} \\
    & & \multicolumn{1}{l|}{Chicago in 1881.} \\
    \hline
    Semantics & Same & Different \\
    \hline
    Model's & Considers the two & Considers the two \\
    Mistake & to be different & to be the same \\
    \hline
    Model & Overly & Overly \\
    Weakness & sensitive & stable \\
    \hline
  \end{tabular}
\caption{
  Adversarial examples in computer vision
  exploit model oversensitivity to small perturbations.
  In contrast, our adversarial examples work because
  models do not realize that a small perturbation
  can completely change the meaning of a sentence.
Images from \citet{szegedy2014intriguing}.}
\label{tab:stability}
\end{table}
    
Instead of relying on paraphrasing,
we use perturbations that do alter semantics
to build \emph{concatenative} adversaries,
which generate examples of the form $(p + s, q, a)$ for some sentence $s$.
In other words, concatenative adversaries
add a new sentence to the end of the paragraph,
and leave the question and answer unchanged.
Valid adversarial examples are precisely those
for which $s$ does not contradict the correct answer;
we refer to such sentences as being \emph{compatible} with $(p, q, a)$.
We use semantics-altering perturbations to that ensure that $s$ is compatible,
even though it may have many words in common with the question $q$.
Existing models are bad at distinguishing these sentences from sentences
that do in fact address the question,
indicating that they suffer not from oversensitivity
but from \emph{overstability} to semantics-altering edits.
Table~\ref{tab:stability} summarizes this important distinction.

The decision to always append $s$ to the end of $p$ 
is somewhat arbitrary;
we could also prepend it to the beginning,
though this would violate the expectation of the first sentence
being a topic sentence.
Both are more likely to preserve the validity of the example than 
inserting $s$ in the middle of $p$,
which runs the risk of breaking coreference links.

\begin{figure*}[t] 
\small
\begin{center} 
  \includegraphics[scale=0.82]{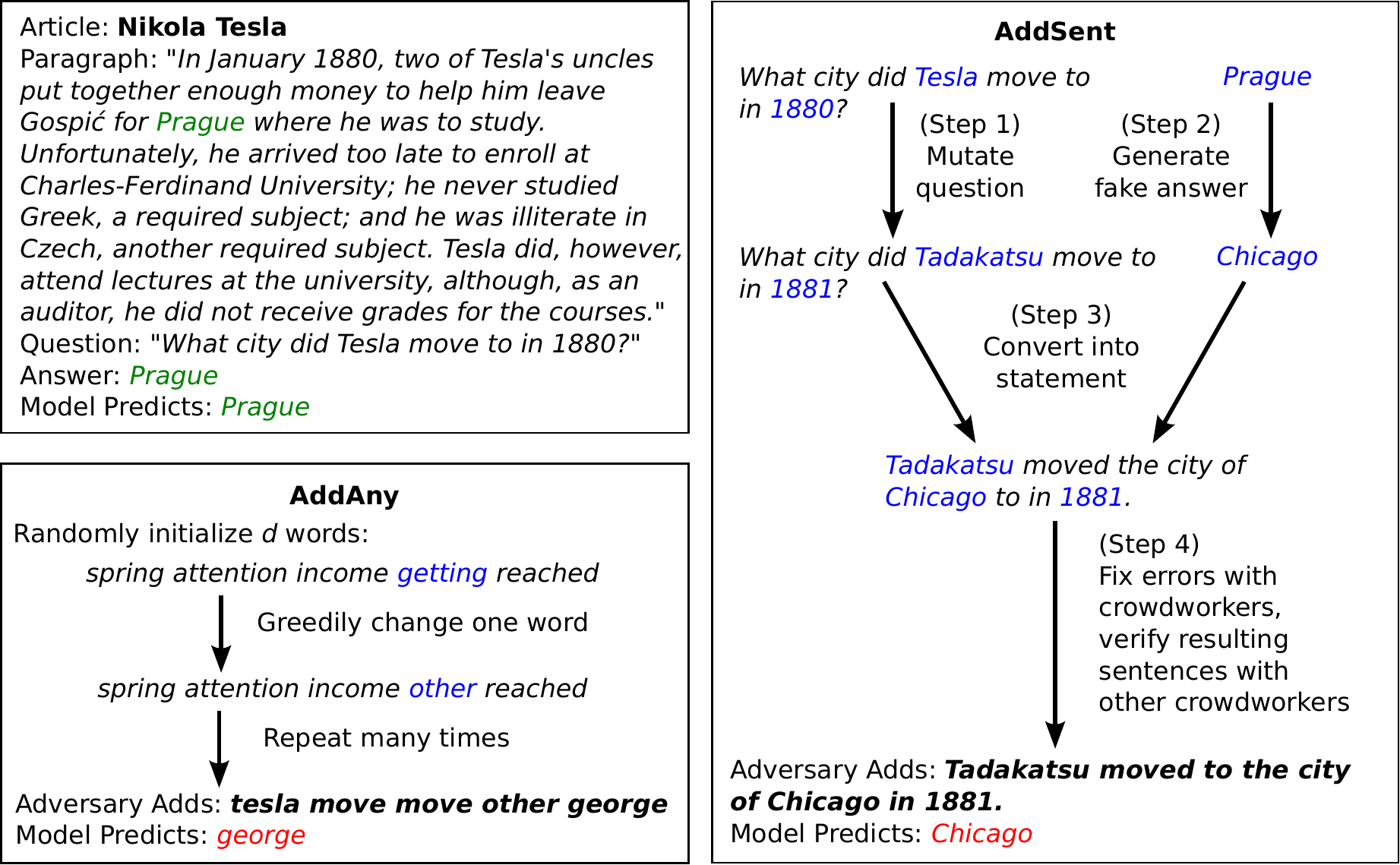}
\end{center} 
\caption{An illustration of the \addsent and \addwords adversaries.}
\label{fig:methods}
\end{figure*}

Now, we describe two concrete concatenative adversaries, as well as two variants.
\addsent, our main adversary, adds grammatical sentences that look similar to the question.
In contrast, \addwords adds arbitrary sequences of English words,
giving it more power to confuse models.
Figure~\ref{fig:methods} illustrates these two main adversaries.

\subsubsection{\addsent}
\addsent uses a four-step procedure
to generate sentences that look similar to the question,
but do not actually contradict the correct answer.
Refer to Figure~\ref{fig:methods} for an illustration of these steps.

In Step 1, we apply semantics-altering perturbations to the question,
in order to guarantee that the resulting adversarial sentence is compatible.
We replace nouns and adjectives with antonyms from
WordNet \citep{fellbaum1998wordnet},
and change named entities and numbers to 
the nearest word in GloVe word vector space\footnote{
We use 100-dimensional GloVe vectors trained on Wikipedia
and Euclidean distance to define nearby words.}
\cite{pennington2014glove}
with the same part of speech.\footnote{
  We choose the nearest word whose most common gold POS tag
in the Penn Treebank \citep{marcus1999ptb}
matches the predicted POS tag of the original word, according to CoreNLP.
If none of the nearest $100$ words satisfy this,
we just return the single closest word.}
If no words are changed during this step,
the adversary gives up and immediately returns the original example.
For example, given the question 
\nl{What ABC division handles domestic television distribution?},
we would change \nl{ABC} to \nl{NBC} (a nearby word in vector space)
and \nl{domestic} to \nl{foreign} (a WordNet antonym),
resulting in the question, 
\nl{What NBC division handles foreign television distribution?}

In Step 2, we create a fake answer that has the same ``type''
as the original answer.
We define a set of $26$ types, corresponding to NER and POS tags from
Stanford CoreNLP \cite{manning2014stanford},
plus a few custom categories (e.g., abbreviations),
and manually associate a fake answer with each type.
Given the original answer to a question, we compute its type and
return the corresponding fake answer.
In our running example, the correct answer was not tagged
as a named entity, and has the POS tag \texttt{NNP},
which corresponds to the fake answer \nl{Central Park.}

In Step 3, we combine the altered question and fake answer into declarative form,
using a set of roughly $50$
manually-defined rules over CoreNLP constituency parses.
For example,
\nl{What ABC division handles domestic television distribution?}
triggers a rule that converts questions of the form
``\textit{what/which} \texttt{NP$_1$} \texttt{VP$_1$} ?''
to 
``The \texttt{NP$_1$} of \texttt{[Answer]} \texttt{VP$_1$}''.
After incorporating the alterations and fake answer from the previous steps,
we generate the sentence,
\nl{The NBC division of Central Park handles foreign television distribution.}

The raw sentences generated by Step 3
can be ungrammatical or otherwise unnatural
due to the incompleteness of our rules and errors in constituency parsing.
Therefore, in Step 4, we fix errors in these sentences via crowdsourcing.
Each sentence is edited independently by five workers on Amazon Mechanical Turk,
resulting in up to five sentences for each raw sentence. 
Three additional crowdworkers then filter out
sentences that are ungrammatical or incompatible,
resulting in a smaller (possibly empty) set of human-approved sentences.
The full \addsent adversary
runs the model $f$ as a black box on every human-approved sentence, and
picks the one that makes the model give the worst answer.
If there are no human-approved sentences, the adversary
simply returns the original example.

\textbf{A model-independent adversary}.
\addsent requires a small number of queries
to the model under evaluation.
To explore the possibility of an adversary that is
completely model-independent, we also introduce
\addonesent, which adds a random human-approved sentence to the paragraph.
In contrast with prior work in computer vision
\citep{papernot2017blackbox,narodytska2016blackbox,moosavidezfooli2017universal},
\addonesent does not require any access to the model
or to any training data: it generates adversarial examples
based solely on the intuition that existing models are overly stable.

\subsubsection{\addwords}
For \addwords, the goal is to choose any sequence of $d$ words,
regardless of grammaticality.
We use local search to adversarially choose a distracting sentence $s = w_1\, w_2 \dots w_d$.
Figure~\ref{fig:methods} shows an example of \addwords with $d=5$ words;
in our experiments, we use $d =10$.

We first initialize words $w_1, \dotsc, w_d$
randomly from a list of common English words.\footnote{
  We define common words as the $1000$ most frequent words in the
Brown corpus \citep{francis1979brown}.}
Then, we run $6$ epochs of local search,
each of which iterates over the indices $i \in \{1, \dotsc, d\}$ in a random order.
For each $i$, we randomly generate a set of candidate words $W$ as
the union of $20$ randomly sampled common words
and all words in $q$.
For each $x \in W$, we generate the sentence with
$x$ in the $i$-th position and $w_j$ in the $j$-th position for each $j \ne i$.
We try adding each sentence to the paragraph and
query the model for its predicted probability distribution over answers.
We update $w_i$ to be the $x$ that minimizes the
expected value of the F1 score over the model's output distribution.
We return immediately if the model's argmax prediction has $0$ F1 score.
If we do not stop after $3$ epochs,
we randomly initialize $4$ additional word sequences,
and search over all of these random initializations in parallel.

\addwords requires significantly more model access than \addsent:
not only does it query the model many times during the search process,
but it also assumes that the model returns a probability distribution
over answers, instead of just a single prediction.
Without this assumption, we would have to rely on something like
the F1 score of the argmax prediction,
which is piecewise constant and therefore harder to optimize.
``Probabilistic'' query access is still weaker than
access to gradients,
as is common in computer vision
\cite{szegedy2014intriguing,goodfellow2015explaining}.

We do not do anything to ensure that the sentences
generated by this search procedure do not contradict the original answer.
In practice, the generated ``sentences'' are
gibberish that use many question words but
have no semantic content (see Figure~\ref{fig:methods} for an example).

Finally, we note that both \addsent and \addwords
try to incorporate words from the question
into their adversarial sentences.
While this is an obvious way to draw the model's attention,
we were curious if we could also distract the model without
such a straightforward approach.
To this end, we introduce a variant of \addwords called
\addcommon, which is exactly like \addwords except
it only adds common words.

\section{Experiments}
\subsection{Setup}
For all experiments, 
we measure adversarial F1 score \cite{rajpurkar2016squad}
across $1000$ randomly sampled examples from the
SQuAD development set (the test set is not publicly available).
Downsampling was helpful because \addwords and \addcommon 
can issue thousands of model queries per example,
making them very slow.
As the effect sizes we measure are large,
this downsampling does not hurt statistical significance.

\subsection{Main Experiments}
\begin{table}
  \centering
  \small
  \begin{tabular}{|l|cccc|}
    \hline
    & Match & Match & BiDAF & BiDAF \\
    & Single & Ens. & Single & Ens. \\
    \hline
    Original    & $71.4$ & $75.4$ & $75.5$ & $80.0$ \\
    \hline
    \addsent    & $27.3$ & $29.4$ & $34.3$ & $34.2$ \\
    \addonesent & $39.0$ & $41.8$ & $45.7$ & $46.9$ \\
    \hline
    \addwords   & { }$7.6$ & $11.7$ & { }$4.8$ & { }$2.7$ \\
    \addcommon  & $38.9$ & $51.0$ & $41.7$ & $52.6$ \\
    \hline
  \end{tabular}
  \caption{Adversarial evaluation on the Match-LSTM 
    and BiDAF systems.
    All four systems can be fooled by adversarial examples.}
  \label{tab:experiments}
\end{table}
\begin{table}
  \centering
  \small
  \begin{tabular}{|l|ccc|}
    \hline
    Model                 & Original & \addsent & \addonesent \\
    \hline                           
    ReasoNet-E            & $\bf 81.1$   & $39.4$   & $49.8$ \\
    SEDT-E                & $80.1$   & $35.0$   & $46.5$ \\
    BiDAF-E               & $80.0$   & $34.2$   & $46.9$ \\
    Mnemonic-E            & $79.1$   & $\bf 46.2$   & $\bf 55.3$ \\
    Ruminating            & $78.8$   & $37.4$   & $47.7$ \\
    jNet                  & $78.6$   & $37.9$   & $47.0$ \\
    Mnemonic-S            & $78.5$   & $\bf 46.6$   & $\bf 56.0$ \\
    ReasoNet-S            & $78.2$   & $39.4$   & $50.3$ \\
    MPCM-S                & $77.0$   & $40.3$   & $50.0$ \\
    SEDT-S                & $76.9$   & $33.9$   & $44.8$ \\
    RaSOR                 & $76.2$   & $39.5$   & $49.5$ \\
    BiDAF-S               & $75.5$   & $34.3$   & $45.7$ \\
    Match-E               & $75.4$   & $29.4$   & $41.8$ \\
    Match-S               & $71.4$   & $27.3$   & $39.0$ \\
    DCR                   & $69.3$   & $37.8$   & $45.1$ \\
    Logistic              & $50.4$   & $23.2$   & $30.4$ \\
    \hline
  \end{tabular}
  \caption{\addsent and \addonesent on all sixteen models, 
  sorted by F1 score the original examples. 
  S = single, E = ensemble.}
  \label{tab:other-models}
\end{table}
Table~\ref{tab:experiments} shows the performance of the Match-LSTM and BiDAF models
against all four adversaries.
Each model incurred a significant
accuracy drop under every form of adversarial evaluation.
\addsent made average F1 score across
the four models fall from $75.7\%$ to $31.3\%$.
\addwords was even more effective,
making average F1 score fall to $6.7\%$.
\addonesent retained much of the effectiveness of \addsent,
despite being model-independent.
Finally, \addcommon caused average F1 score to fall to $46.1\%$,
despite only adding common words.

We also verified that our adversaries were general enough
to fool models that we did not use during development.
We ran \addsent on twelve published models
for which we found publicly available test-time code;
we did not run \addwords on these models, 
as not all models exposed output distributions.
As seen in Table~\ref{tab:other-models},
no model was robust to adversarial evaluation;
across the sixteen total models tested, average F1 score
fell from $75.4\%$ to $36.4\%$ under \addsent.

It is noteworthy that the Mnemonic Reader models \cite{hu2017mnemonic} 
outperform the other models by about $6$ F1 points.
We hypothesize that Mnemonic Reader's self-alignment layer,
which helps model long-distance relationships between 
parts of the paragraph, makes it better at locating all pieces of evidence that
support the correct answer. 
Therefore, it can be more confident in the correct answer,
compared to the fake answer inserted by the adversary.

\subsection{Human Evaluation}
\begin{table}
  \centering
  \small
  \begin{tabular}{|l|c|}
    \hline
    & Human \\
    \hline
    Original    & $92.6$ \\
    \addsent    & $79.5$ \\
    \addonesent & $89.2$ \\
    \hline
  \end{tabular}
  \caption{Human evaulation on adversarial examples.
    Human accuracy drops on \addsent mostly due
    to unrelated errors;
    the \addonesent numbers show that humans are robust
    to adversarial sentences.}
  \label{tab:human}
\end{table}
To ensure our results are valid,
we verified that humans are not also fooled by our adversarial examples.
As \addwords requires too many model queries to run against humans, 
we focused on \addsent.
We presented each original and adversarial paragraph-question pair
to three crowdworkers, and asked them to select the correct answer
by copy-and-pasting from the paragraph.
We then took a majority vote over the three responses
(if all three responses were different, we picked one at random).
These results are shown in Table~\ref{tab:human}.
On original examples, our humans are actually slightly better
than the reported number of $91.2$ F1 on the entire development set.
On \addsent, human accuracy drops by $13.1$ F1 points,
much less than the computer systems.

Moreover, much of this decrease can be explained by
mistakes unrelated to our adversarial sentences.
Recall that \addsent picks the worst case over
up to five different paragraph-question pairs.
Even if we showed the same original example to five sets of three crowdworkers,
chances are that at least one of the five groups would make a mistake,
just because humans naturally err.
Therefore, it is more meaningful to evaluate humans on \addonesent,
on which their accuracy drops by only $3.4$ F1 points.

\subsection{Analysis}
Next, we sought to better understand the behavior
of our four main models under adversarial evaluation.
To highlight errors caused by the adversary,
we focused on examples where the model
originally predicted the (exact) correct answer.
We divided this set into 
``model successes''---examples
where the model continued being correct during adversarial
evaluation---and ``model failures''---examples 
where the model gave a wrong answer during adversarial evaluation.

\subsubsection{Manual verification} 
First, we verified that the sentences added by 
\addsent are actually grammatical and compatible.
We manually checked $100$ randomly chosen BiDAF Ensemble failures.
We found only one where the sentence could be interpreted
as answering the question:
in this case, \addsent replaced the word \nl{Muslim}
with the related word \nl{Islamic}, so
the resulting adversarial sentence still contradicted the correct answer.
Additionally, we found $7$ minor grammar errors, such as 
subject-verb disagreement 
(e.g., \nl{The Alaskan Archipelago \textbf{are} made up almost entirely of hamsters.})
and misuse of function words
(e.g., \nl{The gas \textbf{of} nitrogen makes up 21.8 \% of \textbf{the} Mars's atmosphere.}),
but no errors that materially impeded understanding of the sentence.

We also verified compatibility for \addwords.
We found no violations out of $100$ randomly chosen BiDAF Ensemble failures.

\subsubsection{Error analysis}
Next, we wanted to understand what types of errors
the models made on the \addsent examples.
In $96.6\%$ of model failures,
the model predicted a span in the adversarial sentence.
The lengths of the predicted answers were mostly similar 
to those of correct answers,
but the BiDAF models occasionally predicted very long spans.
The BiDAF Single model predicted an answer of more than $29$ words---the
length of the longest answer in the SQuAD development set---on
$5.0\%$ of model failures;
for BiDAF Ensemble, this number was $1.6\%$.
Since the BiDAF models independently predict the start and end positions of the answer,
they can predict very long spans when the
end pointer is influenced by the adversarial sentence,
but the start pointer is not.
Match-LSTM has a similar structure,
but also has a hard-coded rule that stops
it from predicting very long answers.

We also analyzed human failures---examples where the humans
were correct originally, but wrong during adversarial evaluation.
Humans predicted from the adversarial sentence
on only $27.3\%$ of these error cases, which confirms that
many errors are normal mistakes unrelated to adversarial sentences.

\subsubsection{Categorizing \addsent sentences}
We then manually examined sentences generated by \addsent.
In $100$ BiDAF Ensemble failures,
we found $75$ cases where an entity name was changed
in the adversarial sentence,
$17$ cases where numbers or dates were changed, and $33$
cases where an antonym of a question word was used.\footnote{
These numbers add up to more than $100$ because
more than one word can be altered per example.}
Additionally, $7$ sentences had other miscellaneous perturbations
made by crowdworkers during Step 4 of \addsent.
For example, on a question about the \nl{Kalven Report},
the adversarial sentence discussed \nl{The statement Kalven cited} instead;
in another case, the question, \nl{How does Kenya curb corruption?}
was met by the unhelpful sentence, \nl{Tanzania is curbing corruption}
(the model simply answered, \nl{corruption}).

\subsubsection{Reasons for model successes}
\begin{figure}[t] 
\begin{center} 
  \vspace{-0.30in}
  \includegraphics[scale=0.40]{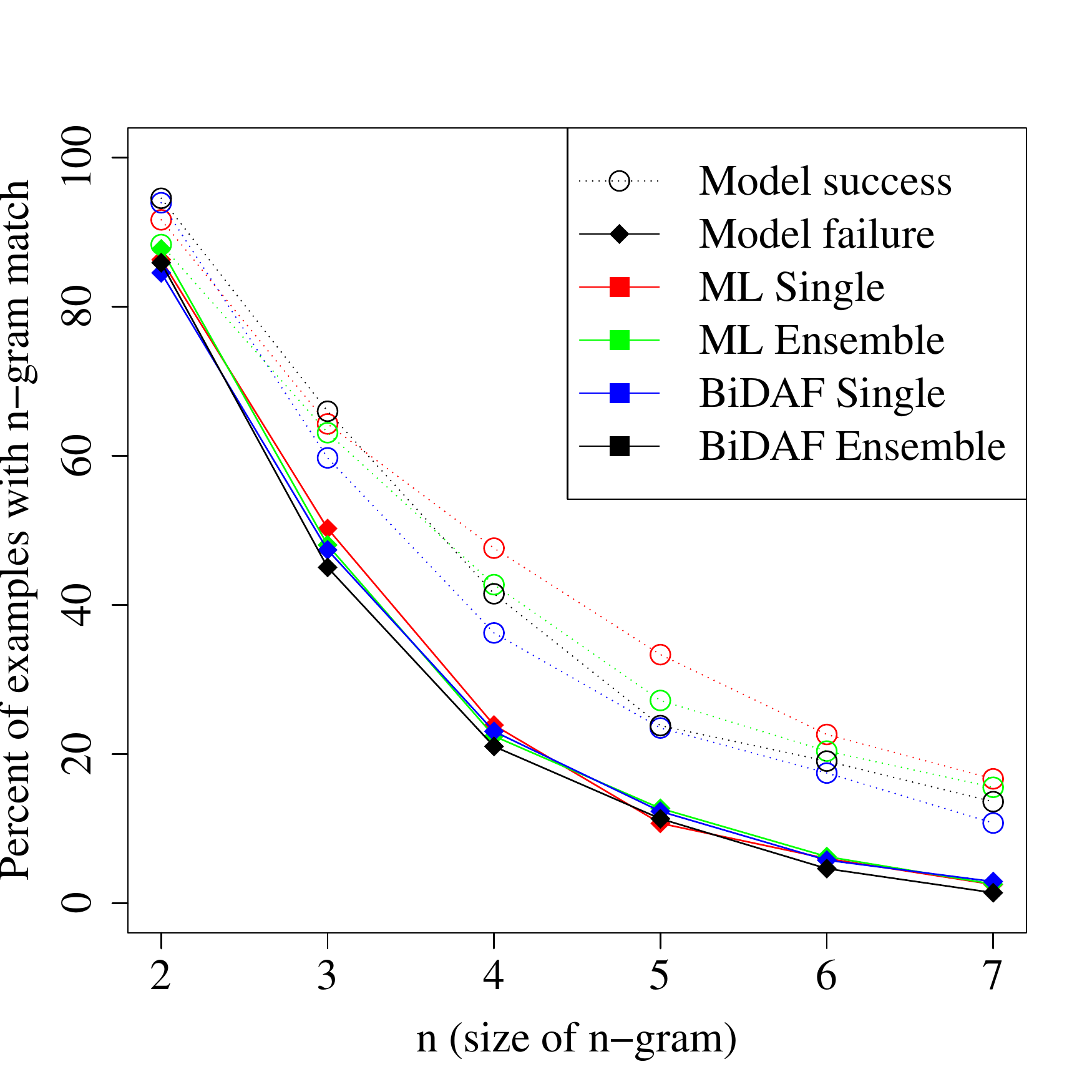}
  \vspace{-0.20in}
\end{center} 
\caption{Fraction of model successes and failures 
  on \addsent for which the question has an exact $n$-gram
  match with the original paragraph.  
  For each model and each value of $n$, 
  successes are more likely to have an
  $n$-gram match than failures.
  }
\label{fig:ngram}
\end{figure}
\begin{figure}[t] 
\begin{center} 
  \vspace{-0.30in}
  \includegraphics[scale=0.40]{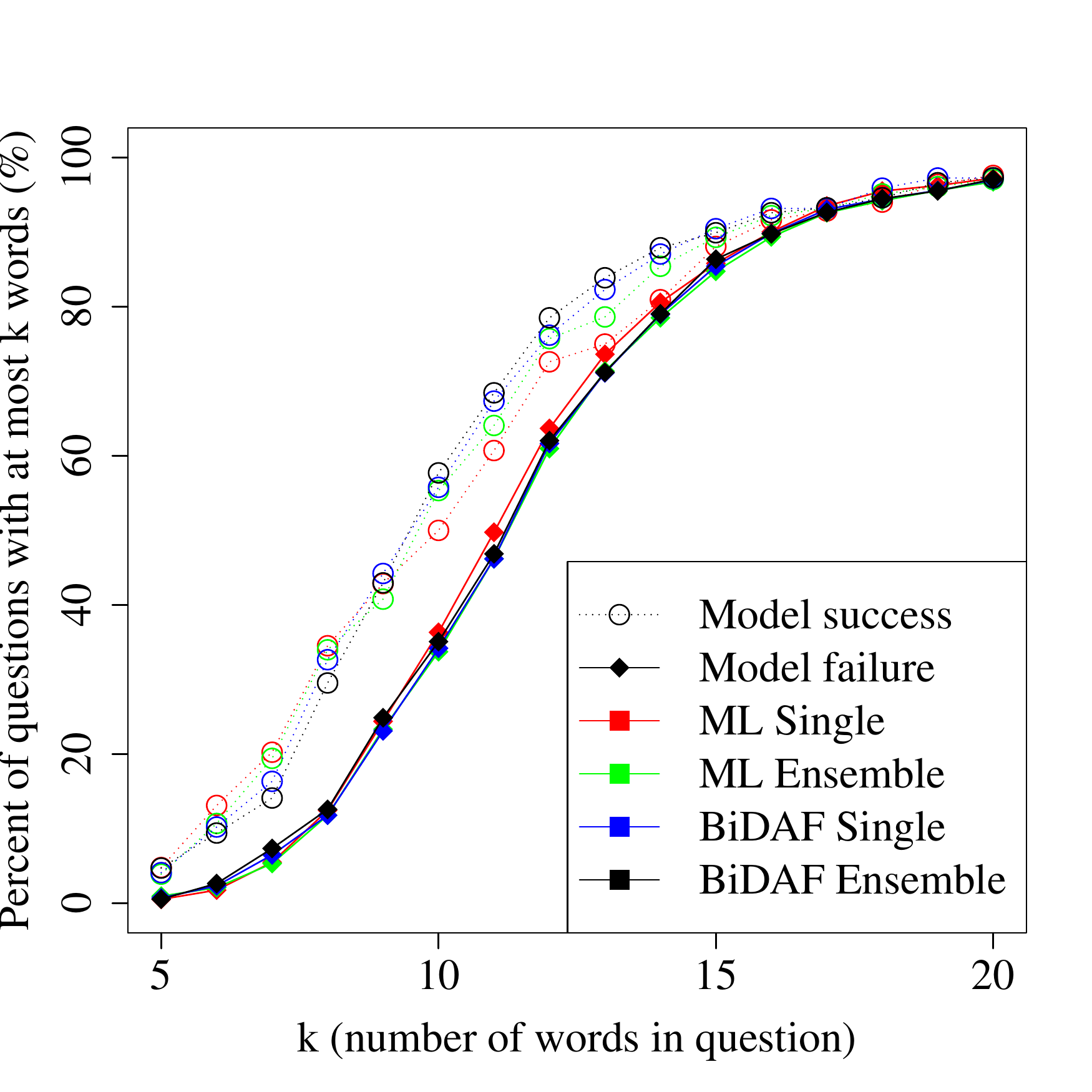}
  \vspace{-0.20in}
\end{center} 
\caption{
  For model successes and failures on \addsent,
  the cumulative distribution function of 
  the number of words in the question
  (for each $k$, what fraction of questions have $\le k$ words).
  Successes are more likely to involve short questions.
}
\label{fig:q-len}
\end{figure}
Finally, we sought to understand the factors
that influence whether the model will be robust to 
adversarial perturbations on a particular example.
First, we found that models do well
when the question has an exact $n$-gram match
with the original paragraph.
Figure~\ref{fig:ngram} plots the fraction of examples for
which an $n$-gram in the question appears verbatim
in the original passage; this is much higher for
model successes.
For example, $41.5\%$ of BiDAF Ensemble successes
had a $4$-gram in common with the original paragraph,
compared to only $21.0\%$ of model failures.

We also found that models succeeded more often on short questions.
Figure~\ref{fig:q-len} shows the distribution of
question length on model successes and failures;
successes tend to involve shorter questions.
For example, 
$32.7\%$ of the questions in BiDAF Ensemble successes
were $8$ words or shorter, compared to only $11.8\%$
for model failures.
This effect arises because \addsent 
always changes at least one word in the question.
For long questions, changing one word leaves many others unchanged, 
so the adversarial sentence still has many words in common with the question.
For short questions, changing one content word may be enough to make
the adversarial sentence completely irrelevant.

\subsection{Transferability across Models}
\begin{table}[t]
  \centering
  \small
  \begin{tabular}{|l|cccc|}
    \hline
    & \multicolumn{4}{c|}{Model under Evaluation} \\
    \multirow{2}{*}{Targeted Model} & ML     & ML   & BiDAF  & BiDAF \\
                                    & Single & Ens. & Single & Ens.  \\
    \hline
    \textbf{\addsent} & & & & \\
    ML Single                       & $27.3$ & $33.4$ & $40.3$ & $39.1$ \\
    ML Ens.                         & $31.6$ & $29.4$ & $40.2$ & $38.7$ \\
    BiDAF Single                    & $32.7$ & $34.8$ & $34.3$ & $37.4$ \\
    BiDAF Ens.                      & $32.7$ & $34.2$ & $38.3$ & $34.2$ \\
    \hline
    \textbf{\addwords} & & & & \\
    ML Single                       & { }$7.6$ & $54.1$ & $57.1$ & $60.9$ \\
    ML Ens.                         & $44.9$ & $11.7$ & $50.4$ & $54.8$ \\
    BiDAF Single                    & $58.4$ & $60.5$ & { }$4.8$ & $46.4$ \\
    BiDAF Ens.                      & $48.8$ & $51.1$ & $25.0$ & { }$2.7$ \\
    \hline
  \end{tabular}
  \caption{Transferability of adversarial examples across models.
    Each row measures performance on adversarial examples
    generated to target one particular model;
    each column evaluates one (possibly different) model on these examples.
  }
  \label{tab:transferability}
\end{table}
In computer vision,
adversarial examples that fool one model also tend to fool other models
\cite{szegedy2014intriguing,moosavidezfooli2017universal};
we investigate whether the same pattern holds for us.
Examples from \addonesent clearly do transfer across models,
since \addonesent always adds the same adversarial sentence regardless of model.

Table~\ref{tab:transferability} shows the results of evaluating
the four main models on adversarial examples generated by running
either \addsent or \addwords against each model.
\addsent adversarial examples transfer between models quite effectively;
in particular, they are harder than \addonesent examples,
which implies that examples that fool one model
are more likely to fool other models.
The \addwords adversarial examples exhibited more limited transferability
between models. 
For both \addsent and \addwords, examples transferred slightly better
between single and ensemble versions of the same model.

\subsection{Training on Adversarial Examples}
\begin{table}[t]
  \centering
  \small
  \begin{tabular}{|l|cc|}
    \hline
    & \multicolumn{2}{c|}{Training data} \\
    Test data & Original & Augmented \\
    \hline
    Original & $75.8$ & $75.1$ \\
    \addsent & $34.8$ & $70.4$ \\
    \addmod  & $34.3$ & $39.2$ \\
    \hline
  \end{tabular}
  \caption{Effect of training the BiDAF Single model
    on the original training data alone (first column)
  versus augmenting the data with raw \addsent examples (second column).}
  \label{tab:retraining}
\end{table}

Finally, we tried training on adversarial examples,
to see if existing models can learn to become more robust.
Due to the prohibitive cost of running \addsent or \addwords 
on the entire training set, we instead ran only 
Steps 1-3 of \addsent (everything except crowdsourcing)
to generate a raw
adversarial sentence for each training example.
We then trained the BiDAF model
from scratch on the union of these examples and the
original training data.
As a control, we also trained a second BiDAF model 
on the original training data alone.\footnote{
All previous experiments used parameters released by
\citet{seo2016bidaf}}

The results of evaluating these models are shown in Table~\ref{tab:retraining}.
At first glance, training on adversarial data seems effective,
as it largely protects against \addsent.
However, further investigation shows that training on these examples
has only limited utility.
To demonstrate this, we created a variant of \addsent called
\addmod, which differs from \addsent in two ways:
it uses a different set of fake answers
(e.g., \texttt{PERSON} named entities map to \nl{Charles Babbage} instead of \nl{Jeff Dean}),
and it prepends the adversarial sentence to the beginning of the paragraph
instead of appending it to the end.
The retrained model does almost as badly as the original one
on \addmod, suggesting that it has just learned to ignore the
last sentence and reject the fake answers that \addsent
usually proposed.
In order for training on adversarial examples to actually improve the model,
more care must be taken to ensure that the
model cannot overfit the adversary.

\section{Discussion and Related Work}
Despite appearing successful by standard evaluation metrics,
existing machine learning systems for reading comprehension
perform poorly under adversarial evaluation.
Standard evaluation is overly lenient on 
models that rely on superficial cues.
In contrast, adversarial evaluation reveals that
existing models are overly stable to perturbations
that alter semantics.

To optimize adversarial evaluation metrics,
we may need new strategies for training models.
For certain classes of models and adversaries,
efficient training strategies exist:
for example,
\citet{globerson2006nightmare} 
train classifiers that are optimally robust to
adversarial feature deletion.
Adversarial training \citep{goodfellow2015explaining} 
can be used for any model trained with stochastic gradient descent,
but it requires generating new adversarial examples at every iteration;
this is feasible for images, where fast gradient-based adversaries
exist, but is infeasible for domains where
only slower adversaries are available.

We contrast \emph{adversarial evaluation}, as studied in this work,
with \emph{generative adversarial models}.
While related in name, the two have very different goals.
Generative adversarial models pit a generative model,
whose goal is to generate realistic outputs,
against a discriminative model,
whose goal is to distinguish the generator's outputs from real data
\citep{smith2012adversarial,goodfellow2014gan}. 
\citet{bowman2016continuous} and \citet{li2017adversarial}
used such a setup for sentence and dialogue generation,
respectively.
Our setup also involves a generator and a discriminator
in an adversarial relationship;
however, our discriminative system is tasked with finding
the right answer, not distinguishing the generated examples
from real ones, and our goal is to 
evaluate the discriminative system, not to train the generative one.

While we use adversaries as a way to evaluate language understanding,
robustness to adversarial attacks may also be its own goal
for tasks such as spam detection.
\citet{dalvi2004adversarial} formulated such tasks as a game
between a classifier and an adversary, and analyzed
optimal strategies for each player.
\citet{lowd2005adversarial}
described an efficient attack by which an adversary can 
reverse-engineer the weights of a linear classifier,
in order to then generate adversarial inputs.
In contrast with these methods, we do not make strong structural assumptions
about our classifiers.

Other work has proposed harder test datasets for various tasks.
\citet{levesque2013best} proposed the Winograd Schema challenge,
in which computers must resolve coreference resolution problems
that were handcrafted to require extensive world knowledge.
\citet{paperno2016lambada} constructed the LAMBADA dataset,
which tests the ability of language models to handle long-range dependencies.
Their method relies on the availability of a large initial dataset,
from which they distill a difficult subset;
such initial data may be unavailable for many tasks.
\citet{rimell2009unbounded} showed that dependency parsers 
that seem very accurate by standard metrics perform poorly on 
a subset of the test data that has unbounded dependency constructions.
Such evaluation schemes can only test models on phenomena
that are moderately frequent in the test distribution;
by perturbing test examples, we can introduce out-of-distribution phenomena
while still leveraging prior data collection efforts.

While concatenative adversaries are well-suited to reading comprehension,
other adversarial methods may prove more effective on other tasks.
As discussed previously,
paraphrase generation systems \citep{madnani2010generating}
could be used for adversarial evaluation
on a wide range of language tasks.
Building on our intuition that existing models are overly stable,
we could apply meaning-altering perturbations
to inputs on tasks like machine translation, 
and adversarially choose ones for which the model's
output does \emph{not} change.
We could also adversarially generate new examples
by combining multiple existing ones, in the spirit of
Data Recombination \citep{jia2016recombination}.
The Build It, Break It shared task \cite{bender2017buildit}
encourages researchers to adversarially design
minimal pairs to fool sentiment analysis and semantic role labeling systems.

Progress on building systems that truly understand language
is only possible if our evaluation metrics
can distinguish real intelligent behavior from shallow pattern matching.
To this end, we have released scripts to run \addsent on any SQuAD system,
as well as code for \addwords.
We hope that our work will motivate
the development of more sophisticated
models that understand language at a deeper level.

\paragraph{Acknowledgments.}
We thank Pranav Rajpurkar for his help with various SQuAD models.
This work was supported by the 
NSF Graduate Research Fellowship under Grant No. DGE-114747,
and funding from Facebook AI Research and Microsoft.

\paragraph{Reproducibility.} All code, data, and experiments for this
paper are available on the CodaLab platform at
{\small \url{https://worksheets.codalab.org/worksheets/0xc86d3ebe69a3427d91f9aaa63f7d1e7d/}}.

%\section*{Acknowledgments}

\bibliographystyle{emnlp_natbib}
\bibliography{refdb/all}

\end{document}